\documentclass[runningheads]{llncs}
\usepackage{graphicx}
\usepackage{booktabs}
\usepackage{multirow}
\usepackage{amsmath}

\usepackage{bm}
\usepackage[utf8]{inputenc}
\usepackage{amssymb, latexsym}
\usepackage{tikz}
\usetikzlibrary{calc, decorations.pathreplacing}
\usetikzlibrary{fadings}
\usepackage{color, xcolor}
\usepackage{pgfplots}
\usepackage{pgf-umlsd}
\usepackage{ifthen}
\usetikzlibrary{fadings}
\usetikzlibrary{arrows,decorations.markings}
\usetikzlibrary{spy}
\usepackage{mathrsfs}

\usepackage{floatrow}
\usepackage{caption} 
\usepackage{subcaption}
\usepackage[framemethod=tikz]{mdframed}

\usepackage{makecell}

\usepackage{todonotes}

\newcolumntype{P}[1]{>{\centering\arraybackslash}p{#1}}

\newcommand{\Rho}{\mathrm{P}}
\begin{document}

\title{Deep Modeling of Growth Trajectories for Longitudinal Prediction of Missing Infant Cortical Surfaces
}
\titlerunning{Deep Longitudinal Modeling of Infant Cortical Surfaces}
\author{Peirong Liu\inst{1} \and Zhengwang Wu \inst{2} \and Gang Li \inst{2} \and
Pew-Thian Yap\inst{2} \and Dinggang~Shen\inst{1,2}
}
\authorrunning{Liu et al.}
%
\institute{Department of Computer Science, University of North Carolina at Chapel Hill, NC, USA
\and
Department of Radiology and BRIC, University of North Carolina at Chapel Hill, NC, USA
}
\maketitle              
\begin{abstract}
Charting cortical growth trajectories is of paramount importance for understanding brain development. However, such analysis necessitates the collection of longitudinal data, which can be challenging due to subject dropouts and failed scans. In this paper, we will introduce a method for longitudinal prediction of cortical surfaces using a spatial graph convolutional neural network (GCNN), which extends conventional CNNs from Euclidean to curved manifolds. The proposed method is designed to model the cortical growth trajectories and jointly predict inner and outer cortical surfaces at multiple time points. Adopting a binary flag in loss calculation to deal with missing data, we fully utilize all available cortical surfaces for training our deep learning model, without requiring a complete collection of longitudinal data. Predicting the surfaces directly allows cortical attributes such as cortical thickness, curvature, and convexity to be computed for subsequent analysis.
We will demonstrate with experimental results that our method is capable of capturing the nonlinearity of spatiotemporal cortical growth patterns and can predict cortical surfaces with improved accuracy.

\keywords{Graph Convolutional Neural Networks \and Infant Cortical Surfaces \and Longitudinal Prediction \and Shape Analysis \and Missing Data}
\end{abstract}

\section{Introduction}
\label{intro}
Temporal mapping of cortical changes is crucial for understanding normal and abnormal brain development. 
However, such analysis requires longitudinal followup scans, which can be challenging to acquire due to subject dropouts and failed scans. The easiest way to deal with missing data is by discarding the data of subjects with missing scans. Though convenient, this approach discards a huge amount of useful information and leaves a smaller subset of data for analysis with reduced statistical sensitivity. To make full use of available data, we introduce in this paper a deep learning approach to predicting missing surfaces. 




Meng et al.~\cite{meng2016rf} recently employed random forest for longitudinal prediction of infant cortical thickness in an incomplete dataset. They first imputed the missing data and then used the imputed dataset to train their prediction model. While effective, their method is limited in the following ways: (i) Cortical surfaces need to be mapped onto a common sphere, which is a time consuming process that can lose surface topological information; (ii) Prediction is limited to cortical attributes, such as thickness; the actual surfaces are not generated, (iii) Cortical attributes are predicted at each time point independently, disregarding temporal consistency, and (iv)  The imputed surfaces, when used for training, can introduce errors.
Another attempt to predict cortical surfaces was carried out by Rekik et al. \cite{rekik2015}, where they proposed a learning-based framework for predicting dynamic postnatal changes in the cortical shape based on the cortical surfaces at birth using varifold metric for surface regression. Their method however requires full longitudinal scans, which are not always available.



To address the above-mentioned limitations, we propose in this paper a method for the longitudinal prediction of cortical surfaces based on a spatial graph convolutional neural network (GCNN). 
We first parametrize cortical surfaces spatially using intrinsic local coordinate systems. This allows us to implement an effective means of surface convolution. Such convolution mechanism is incorporated in the graph convolution layers of a dual-channel network that caters to the prediction of both inner and outer cortical surfaces, with vertex-wise cortical thickness as constraint. Longitudinal consistency is enforced by ensuring that the predicted surfaces at not only the final time point, but also the intermediate time points, match actual brain surfaces in the training set. We further adopt a binary flagging mechanism to ensure that the loss associated with a nonexistent surface is not contributing to back-propagation.
Our network is hence flexible and does not require complete longitudinal data for training. Each available cortical surface can contribute to the learning of the growth trajectories for prediction purposes. Experimental results illustrate that the proposed method can accurately predict the non-linear cortical growth with longitudinal consistency. 
The predicted surfaces allow cortical attributes such as cortical thickness to be computed for further analysis. 

\section{Materials}
\label{bg}
This study is approved by the Institutional Review Board of the University of North Carolina (UNC) School of Medicine. 37 healthy infants were recruited and scanned longitudinally at 1, 3, and 6 months of age (see Table~\ref{info}). The acquired T1-weighted (T1w) and T2-weighted (T2w) MR images were processed using the UNC Infant Cortical Surface Pipeline \cite{li2012reconstruction} to obtain the inner and outer surfaces and their vertex-to-vertex correspondences for each hemisphere. The surfaces were then align to a common space using the 4D infant cortical surface atlas described in \cite{wu2017} for cross-sectional and longitudinal correspondences. 
Each cortical surface of left and right hemisphere is represented as a mesh formed by uniform non-intersecting triangles. Mathematically, the mesh can be represented as a graph $\mathcal{G} = (\mathcal{V}, \mathcal{F}$), where $\mathcal{V}$ is the vertex set and $\mathcal{F}\subset\mathcal{V}\times\mathcal{V}\times\mathcal{V}$ is the set of triangular faces, each connecting three vertices. 

\begin{table}[t]
  \renewcommand{\arraystretch}{1.5}
  \caption{Data availability.}
  \centering
  \begin{tabular}{l| c c c}
    \bf{Availability}~ & ~All $1^{\text{st}},3^{\text{rd}},6^{\text{th}}$~ & ~Only $1^{\text{st}},3^{\text{rd}}$~ & ~Only $1^{\text{st}},6^{\text{th}}$~ \\ 
    \hline
    \bf{Number of Subjects}~ & 23 & 5 & 9
  \end{tabular}
  \label{info}
\end{table}
\section{Methods}
\label{3}
In this section, we will first introduce how spatial convolution can be defined on cortical surfaces represented as graphs. Then, we will introduce our GCNN framework for longitudinally consistent prediction of cortical surfaces. 

\subsection{Local Geodesic Polar Grids}
\label{3.1}
Each cortical surface is represented as a triangular mesh $\mathcal{G}=(\mathcal{V},\mathcal{F})$ associated with a vertex-wise feature map $f:\mathcal{V}\mapsto \mathcal{R}$, $f({\mathcal{V}}) = \{f(v):v\in \mathcal{V}\}$. Unlike the various global parametrizations for Euclidean domains, the lack of meaningful global parametrizations for surface meshes \cite{bronstein2017review} forces us to parametrize $\mathcal{G}$ in intrinsic coordinate systems. Following \cite{kokkinos2012shapedesc}, for every vertex $v_i\in\mathcal{V}$, we build a geodesic disc $\mathcal{D}_{v_i} = \{v'|\rho(v',v_i)\leq \rho_{\mathcal{D}}\} \subset\mathcal{V}$ ($\rho_{\mathcal{D}}$ is taken as 1\% of the geodesic diameter of the entire surface mesh \cite{masci2015gcnn}). Then, $\mathcal{D}_{v_i}$ is partitioned into a polar grid with $N_{\rho}$ geodesics bins and $M_\theta$ angular bins, via geodesic outward shooting and unfolding \cite{kimmel1998fm}. Therefore, within the grid $\mathcal{D}_{v_i}$, the local geodesic polar coordinate for any vertex $v_j\in \mathcal{D}_{v_i}$ is ($\rho_{ij},\theta_{ij}$), with $\rho_{ij}$ and $\theta_{ij}$ denoting the geodesic distance and the angular distance between $v_i$ and $v_j$, respectively. For the whole mesh $\mathcal{G}$, this local parameterization can be written as a sparse matrix $({\bm{\Rho}},{\bm{\Theta}})$, with ${\bm{\Rho}} = (\rho_{ij})_{|\mathcal{V}|\times |\mathcal{V}|},~{\bm{\Theta}} = (\theta_{ij})_{|\mathcal{V}|\times |\mathcal{V}|}$, and $i,j = 1,...,|\mathcal{V}|$.

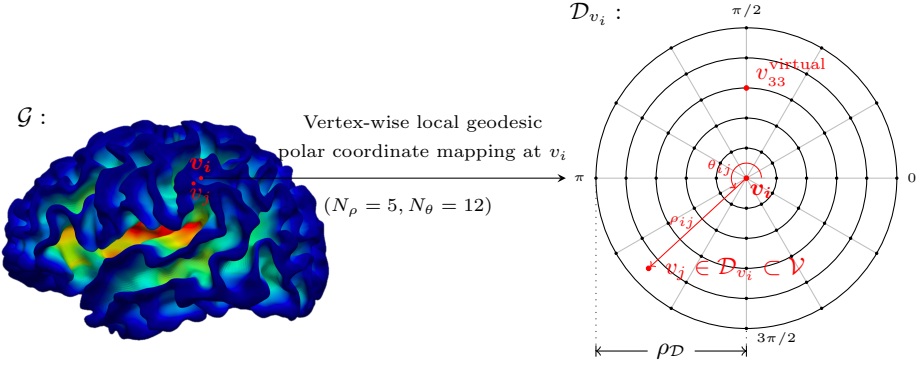
\begin{figure}[t]
\centering
\begin{tikzpicture}
\node[] (input image) at (0,1) {\includegraphics[height=3cm]{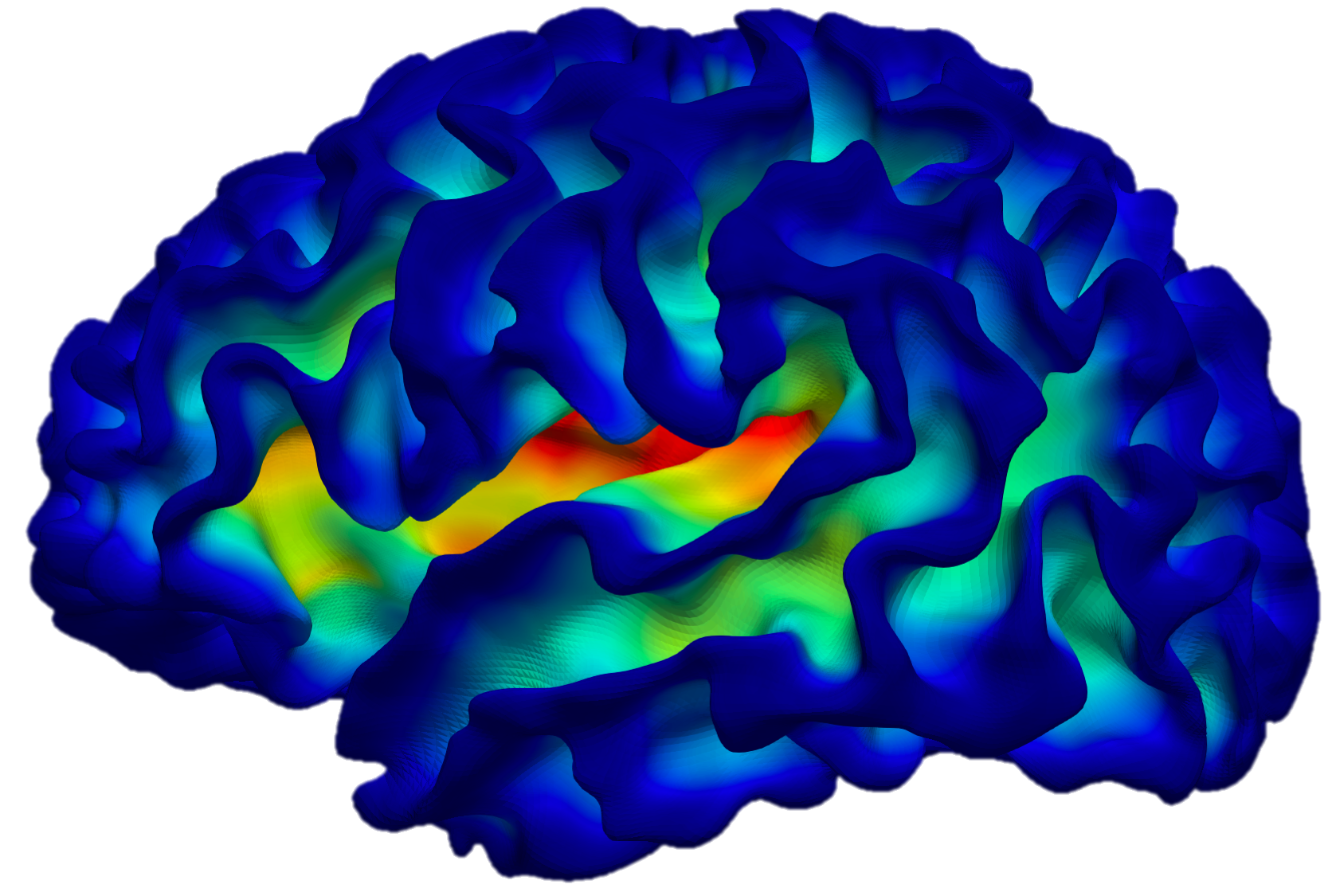}};
\node[] at (-1.75, 2.5) {\begin{tabular}{c}$\mathcal{G}:$
\end{tabular}};
\draw[->,>=stealth, thin] (0.5,1.7) -- (5.35,1.7);
\draw[fill=red, draw=red] (0.5,1.7) circle (0.02cm);
\draw[fill=red, draw=red] (0.4,1.63) circle (0.02cm);
\node[text = red] at (0.5,1.9) {\begin{tabular}{c}${\bm{v_i}}$\end{tabular}};
\node[text = red] at (0.5,1.5) {\begin{tabular}{c}${v_j}$\end{tabular}};
\node at (3.45,2.25) {\begin{tabular}{c}{\scriptsize{Vertex-wise local geodesic}}\\ {\scriptsize{polar coordinate mapping at $v_i$}}\end{tabular}};
\node at (3.25,1.34) {\begin{tabular}{c}{\scriptsize{$(N_{\rho} = 5, N_{\theta} = 12)$}}\end{tabular}};

\foreach \ang in {0,...,11} {
  \draw [lightgray] (7.75,1.7) -- ($(7.75,1.7) + (\ang * 180 / 6:2)$);
}
\foreach \s in {0, 0.4, 0.8, 1.2, 1.6, 2} {
  \draw (7.75,1.7) circle (\s);
  \foreach \ang in {0,...,11} {
   \draw[fill=black, draw=black] ($(7.75,1.7) + (\ang * 180 / 6:\s)$) circle (0.015cm);}
}
\node[text = red] at (7.95,1.55) {\begin{tabular}{c}${\bm{v_i}}$
\end{tabular}};
\draw[->, draw=red, very thin] (7.75,1.7) -- (6.475,0.525);
\node[text = red] at (6.9, 1.2) {\begin{tabular}{c}{\tiny{$\rho_{ij}$}}
\end{tabular}};
\node[text = red] at (7.4, 1.9) {\begin{tabular}{c}{\tiny{$\theta_{ij}$}}
\end{tabular}};
\node[] at (5.75,3.9) {\begin{tabular}{c}${\mathcal{D}_{v_i}}:$
\end{tabular}};

\draw [black!80, dotted] (5.75, -0.6) -- (5.75, 1.7);
\draw [black!80, dotted] (7.75, -0.6) -- (7.75, -0.3);
\draw (5.75, -0.525) -- (5.75, -0.675);
\draw (7.75, -0.525) -- (7.75, -0.675);
\node (A) at (6.75, -0.6) {$\rho_{\mathcal{D}}$};
\draw[->] (A) -- (5.75,-0.6);
\draw[->] (A) -- (7.75,-0.6);

\foreach \ang/\lab/\dir in {
  0/0/right,
  2/{\pi/2}/above,
  4/{\pi}/left} {
  \node [fill=white] at ($(7.75,1.7) + (\ang * 180 / 4:2.025)$) [\dir] {\tiny $\lab$};
}
\node[] at (8.15, -0.4) {\begin{tabular}{c}{\tiny{$3\pi/2$}}
\end{tabular}};
\node[text = red] at (7.6,0.5) {\begin{tabular}{c}$v_j\in \mathcal{D}_{v_i}\subset\mathcal{V}$\end{tabular}};
\draw[->, draw = red, very thin] (7.95,1.7) arc (0:222:0.2);
\draw[lightgray] (7.75,1.7) -- (8,1.7);
\draw[fill=red, draw=red] (7.75,1.7) circle (0.03cm);
\draw[fill=red, draw=red] (6.45,0.5) circle (0.03cm);
\draw[fill=red, draw=red] (7.75,2.9) circle (0.03cm);
\node[text = red] at (8.35,3.1) {\begin{tabular}{c}$v_{_{33}}^{\text{virtual}}$\end{tabular}};
\end{tikzpicture} 

\caption{A local geodesic polar grid $\mathcal{D}_{v_i}$ (5 $\rho$-bins, 12 $\theta$-bins) constructed at vertex $v_i$ on a surface mesh $\mathcal{G}$. $v_j$ is an actual vertex, $v_{_{33}}^{\text{virtual}}$ is a virtual vertex located at the intersection of $3^{rd}$ $\rho$-bin and $3^{rd}$ $\theta$-bin.}
\label{grid}
\end{figure}

\subsection{Spatial Convolution on Cortical Surfaces}
\label{3.2}
\subsubsection{Surface Patch Uniformization.}
\label{3.2.1}
With local parametrization for the whole surface mesh $\mathcal{G}$, local surface patches can be extracted within each geodesic polar grid $D_{v}$ (Sec.~\ref{3.1}). 
However, the distributions of vertices on the patches extracted at different vertices are not necessarily uniform \cite{bronstein2017review}. As in \cite{monti2017monet}, we map surface patches to a common template domain. Specifically, the template contains $N_{\rho}\times N_{\theta}$ virtual vertices $\{v^{\text{virtual}}_{kl}\in\mathcal{D}_v: k = 1,...,N_{\rho}, l = 1,...,N_{\theta}\}$ at the intersections of $N_{\rho}\times N_{\theta}$ bins (Fig.~\ref{grid}). Feature maps $f$ on the virtual vertices can be obtained by weighted interpolation on $\mathcal{D}_v$ based on the actual vertices $\{v'\}\subset\mathcal{D}_v$: 
\begin{equation}
f(\mathcal{D}_v) = \{f(v^{\text{virtual}}_{kl}): k = 1,...,N_{\rho}, l = 1,...,N_{\theta}\},\quad v\in\mathcal{V}
\label{feature}
\end{equation} 
\begin{equation}
f(v^{\text{virtual}}_{kl}) = \sum_{v'\in\mathcal{D}_v}w_{kl}(\bm{u}_v(v'))f(v'),\quad (k,l)\in\{1,...,N_{\rho}\}\times\{1,...,N_{\theta}\}
\end{equation}
where $\bm{u}_v(v') = (\rho{(v,v')},\theta{(v,v')})$ denotes the local geodesic polar coordinate of $v'$ in relation to $v$, $\{w_{kl}(\cdot)\}$ are the parametric Gaussian kernels adopted from \cite{monti2017monet}:
\begin{equation}
w_{kl}({\bm{u}}) = \exp\left\{-\frac{1}{2}(\bm{u}-\bm{u}_{kl})^T{\bm{\sum}}_{kl}^{-1}(\bm{u}-\bm{u}_{kl})\right\}
\label{gaussian}
\end{equation}
where the mean vectors $\{\bm{u}_{kl} \in \mathcal{R}^{2\times 1}\}$ and the diagonal covariance matrices $\{\bm{\sum}_{kl} \in \mathcal{R}^{2\times 2} \}$ are learnable. The parametric kernels with extra degrees of freedom generalize the fixed local Gaussian kernels in \cite{masci2015gcnn}, and naturally solve the origin ambiguity of angular coordinates caused by geodesic polar grids construction in Sec.~\ref{3.1}.

\subsubsection{Spatial Convolution.}
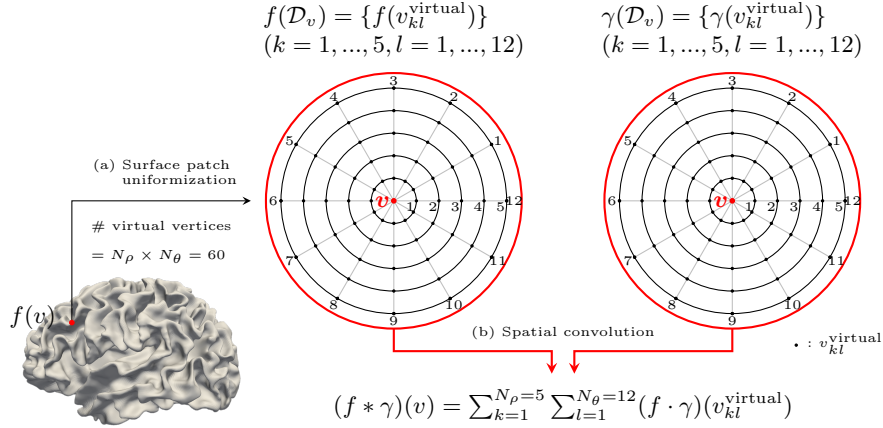
\begin{figure}[t]
\centering
\begin{tikzpicture}
\node[] (input image) at (0,1) {\includegraphics[height=2cm]{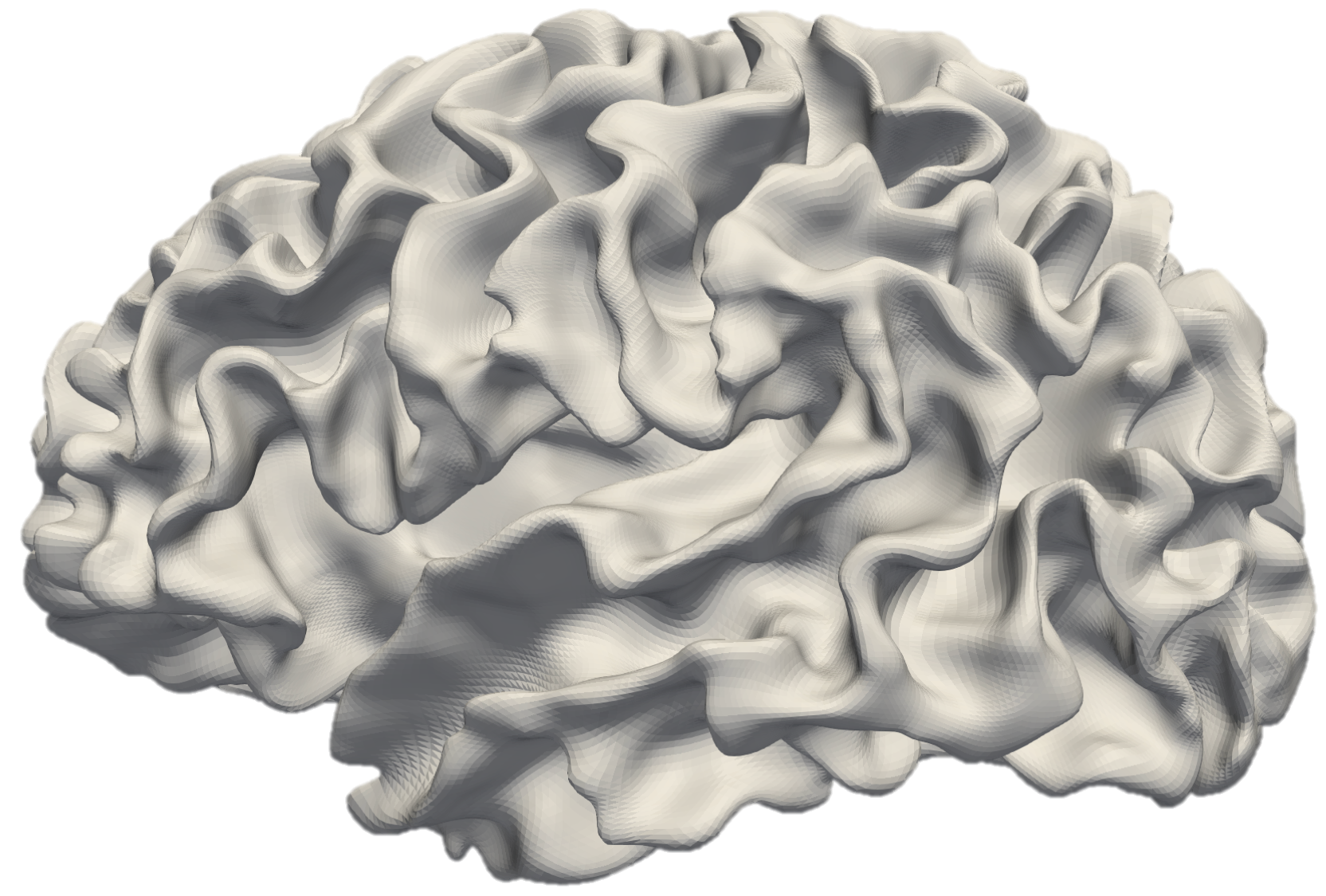}};
\node[] at (3.5, 5.25) {\begin{tabular}{l}$f(\mathcal{D}_v) = \{f(v_{kl}^{\text{virtual}})\}$\\$(k=1,...,5,l=1,...,12)$
\end{tabular}};
\node[] at (8, 5.25) {\begin{tabular}{l}$\gamma(\mathcal{D}_v) = \{\gamma(v_{kl}^{\text{virtual}})\}$\\$(k=1,...,5,l=1,...,12)$
\end{tabular}};
\draw[->,>=stealth, thin] (-0.78,1.38) -- (-0.78,3) -- (1.6,3);
\draw[fill=red, draw=red] (-0.78,1.38) circle (0.03cm);
\node[] at (-1.3, 1.45) {\begin{tabular}{c}$f(v)$\end{tabular}};
\node[] at (0.41,3.5) {\begin{tabular}{c}{\tiny{(a) Surface patch}}\end{tabular}};
\node[] at (0.65,3.3) {\begin{tabular}{c}{\tiny{uniformization}}\end{tabular}};
\node[] at (0.41,2.5) {\begin{tabular}{l}{\tiny{\# virtual vertices}}\\{\tiny{$=N_{\rho}\times N_{\theta}=60$}}\end{tabular}};

\draw [draw = red, thick] (3.5,3) circle (1.7);
\draw [draw = red, thick] (8,3) circle (1.7);

\foreach \ang in {1,...,12} {
  \draw [lightgray] (3.5,3) -- ($(3.5,3) + (\ang * 180 / 6:1.5)$);
  \draw[fill=black, draw=black] ($(3.5,3) + (\ang * 180 / 6:1.5)$) circle (0.015cm);
  \draw[fill=black, draw=black] ($(3.5,3) + (\ang * 180 / 6:1.6)$) node {{\tiny{$\ang$}}};
}

\foreach \ang in {1,...,12} {
  \draw [lightgray] (8,3) -- ($(8,3) + (\ang * 180 / 6:1.5)$);
\draw[fill=black, draw=black] ($(8,3) + (\ang * 180 / 6:1.5)$) circle (0.015cm);
 \draw[fill=black, draw=black] ($(8,3) + (\ang * 180 / 6:1.6)$) node {{\tiny{$\ang$}}};
}
\foreach \s in {0, 0.3, 0.6, 0.9, 1.2, 1.5} {
  \draw (3.5,3) circle (\s);
  \draw (8,3) circle (\s);
  \foreach \ang in {0,...,11} {
  \draw[fill=black, draw=black] ($(3.5,3) + (\ang * 180 / 6:\s)$) circle (0.015cm);
  \draw[fill=black, draw=black] ($(8,3) + (\ang * 180 / 6:\s)$) circle (0.015cm);
  }
}
  
\foreach \s/\d in 
{0.22/1, 0.52/2, 0.82/3, 1.12/4, 1.42/5} {
  \node[] at ($(3.5,3) + (\s ,-0.06)$) {\begin{tabular}{c}{\tiny{\d}}\end{tabular}};
   \node[] at ($(8,3) + (\s ,-0.06)$) {\begin{tabular}{c}{\tiny{\d}}\end{tabular}};
}

\draw [->, > = stealth, draw = red, thick] (3.5,1.3) -- (3.5, 1) --(5.6,1) -- (5.6, 0.7);
\draw [->, > = stealth, draw = red, thick] (8,1.3) -- (8,1) -- (5.9, 1) -- (5.9, 0.7);
\node[] at (5.75,1.28) {\begin{tabular}{c}{\tiny{(b) Spatial convolution}}\end{tabular}};
\node[] at (5.75,0.28) {\begin{tabular}{c}$(f*\gamma)(v) = \sum_{k=1}^{N_{\rho}=5}\sum_{l=1}^{N_{\theta}=12}(f\cdot\gamma)(v^{\text{virtual}}_{kl})$\end{tabular}};

\draw[fill=red, draw=red] (3.5,3) circle (0.03cm);
\node[text = red] at (3.35,3) {\begin{tabular}{c}${\bm{v}}$\end{tabular}};
\draw[fill=red, draw=red] (8,3) circle (0.03cm);
\node[text = red] at (7.85,3) {\begin{tabular}{c}${\bm{v}}$\end{tabular}};
\draw[fill = black] (8.85,1.08) circle (0.015cm);
\node[] at (9.5,1.13) {\begin{tabular}{l}{\tiny{: $v_{kl}^{\text{virtual}}$}}\end{tabular}};

\end{tikzpicture} 
\caption{Spatial graph convolutional layer. Each surface patch extracted by grid $\mathcal{D}_{v}$ has 5 $\rho$-bins and 12 $\theta$-bins, resulting in 60 learnable Gaussian kernels (Eq.~\eqref{gaussian}) for surface patch uniformization. The uniformization step makes it possible to implement the convolution as a scalar product between the patch feature maps and the filter coefficients $\gamma(\mathcal{D}_v)$.
Spatial convolution across the entire triangular mesh $\mathcal{G}$ is similar to Cartesian convolution.}
\label{gcnn}
\end{figure}

With the uniform surface patches, we can define the convolution filter $\bm{\gamma}$ as a real function field on $\mathcal{D}_v$:
\begin{equation}
\gamma(\mathcal{D}_v) = \{\gamma (v^{\text{virtual}}_{kl}): k = 1,...,N_{\rho}, l = 1,...,N_{\theta}\},\quad v\in\mathcal{V}.
\end{equation} 
We define the spatial convolution on cortical surface meshes as 
\begin{equation}
(f*\gamma)(v) = \sum_{k=1}^{N_{\rho}}\sum_{l=1}^{N_{\theta}}(f\cdot\gamma)(v^{\text{virtual}}_{kl}),\quad v^{\text{virtual}}_{kl}\in\mathcal{D}_v,\quad v\in\mathcal{V}
\label{gconv}
\end{equation}
where $\{\gamma(v^{\text{virtual}}_{kl})\}$ are learnable filter coefficients. A spatial graph convolutional layer (Fig.~\ref{gcnn}) consists of two sub-layers that perform (a) Local surface patch uniformization via Eq.~\eqref{feature}; and (b) Spatial convolution via Eq.~\eqref{gconv}.

\subsection{Longitudinal Surface Prediction}
\label{3.3}
Given the baseline ($1^{\text{st}}$ month) inner (`in') and outer (`out') cortical surfaces of an infant subject, i.e., $\{\mathcal{G}^{\text{in}}_1 = (\mathcal{V}^{\text{in}}_1, \mathcal{F}^{\text{in}}_1), \mathcal{G}^{\text{out}}_1 = (\mathcal{V}^{\text{out}}_1, \mathcal{F}^{\text{out}}_1)\}$,
we train a spatial GCNN to predict the inner and outer cortical surfaces at later time points ($3^{\text{rd}}$ month and $6^{\text{th}}$ month).

\subsubsection{Input - Local Parameterization Matrices.}
\label{3.3.1}
Locally parametrized baseline inner and outer cortical surface pair: $I_{\text{G}} = \{\mathcal{G}^{\text{in}}_1 = ({\bm{\Rho}}_1^{\text{in}},{\bm{\Theta}}_1^{\text{in}}),\mathcal{G}^{\text{out}}_1 = ({\bm{\Rho}}_1^{\text{out}},{\bm{\Theta}}_1^{\text{out}})\}$. 

\subsubsection{Input - Local Shape Descriptors.}
Local shape information is captured by computing the geometric relationship between a vertex and its neighbors.
Specifically, we define the vertex-wise neighborhood difference map $I_{\text{ND}}^{\text{in}}(\mathcal{V})$ associated with baseline inner cortical surface $\mathcal{G}^{\text{in}}_1$ as
\begin{equation}
I_{\text{ND}}^{\text{in}}(v)= \{(x^{\text{in}}_1(v) - x^{\text{in}}_1(w), y^{\text{in}}_1(v) - y^{\text{in}}_1(w), z_1^{\text{in}}(v) - z_1^{\text{in}}(w)): w \in \mathcal{N}(v)\}, 
\label{adin}
\end{equation}
where $(x^{\text{in}}_1(v), y^{\text{in}}_1(v), z_1^{\text{in}}(v))$ refers to the $xyz$-coordinates of vertex $v$ on $\mathcal{G}^{\text{in}}_1$ at the $1^{\text{st}}$ month, $\mathcal{N}^{(1)}(v)$ is a set of vertices adjacent to $v$ determined by triangular faces. 
Likewise, the neighborhood difference map of the baseline outer cortical surface $\mathcal{G}^{\text{out}}_1$ is
\begin{equation}
I_{\text{ND}}^{\text{out}}(v)= \{(x^{\text{out}}_1(v) - x^{\text{out}}_1(w), y^{\text{out}}_1(v) - y^{\text{out}}_1(w), z_1^{\text{out}}(v) - z_1^{\text{out}}(w)): w \in \mathcal{N}(v)\}.
\label{adout}
\end{equation}
where $(x^{\text{out}}_1(v), y^{\text{out}}_1(v), z_1^{\text{out}}(v))$ refers to the $xyz$-coordinates of vertex $v$ on $\mathcal{G}^{\text{out}}_1$ at the $1^{\text{st}}$ month.

\subsubsection{Output - Growth Deformations.} 
The output growth maps $O_M(\mathcal{V})$ associated with cortical surface $\mathcal{G}_M$ at the $M$-th month ($M$ = 3, 6) are defined as the vertex-wise displacements of the surface to the $M$-month from the prior time point. More specifically, the growth maps of inner and outer cortical surfaces at the $3^{\text{rd}}$ month are
\begin{equation}
\left\{
\begin{aligned}
&O_3^{\text{in}}(v)=
(x^{\text{in}}_3(v)-x_1^{\text{in}}(v), y^{\text{in}}_3(v)-y_1^{\text{in}}(v), z^{\text{in}}_3(v)-z_1^{\text{in}}(v)), \\
&O_3^{\text{out}}(v)=
(x^{\text{out}}_3(v)-x_1^{\text{out}}(v), y^{\text{out}}_3(v)-y_1^{\text{out}}(v), z^{\text{out}}_3(v)-z_1^{\text{out}}(v)),
\end{aligned}
\right.
\label{ffi1}
\end{equation}
where subscript `3' denotes the $3^{\text{rd}}$ month.
Similarly, the growth maps of inner and outer cortical surfaces at the $6^{\text{th}}$ month are
\begin{equation}
\left\{
\begin{aligned}
&O_6^{\text{in}}(v)=
(x^{\text{in}}_6(v)-x_3^{\text{in}}(v), y^{\text{in}}_6(v)-y_3^{\text{in}}(v), z^{\text{in}}_6(v)-z_3^{\text{in}}(v)), \\
&O_6^{\text{out}}(v)=
(x^{\text{out}}_6(v)-x_3^{\text{out}}(v), y^{\text{out}}_6(v)-y_3^{\text{out}}(v), z^{\text{out}}_6(v)-z_3^{\text{out}}(v)).
\end{aligned}
\right.
\label{ffi2}
\end{equation}
The cortical thickness at vertex $v\in\mathcal{V}$ is computed based on the inner and outer cortical surface pair ($\mathcal{G}_M^{\text{in}}, \mathcal{G}_M^{\text{out}}$) at the $M$-th month as
\begin{equation}
\text{thickness}_M(v) = \Vert(x^{\text{out}}_M(v) - x^{\text{in}}_M(v),y^{\text{out}}_M(v) - y^{\text{in}}_M(v),z^{\text{out}}_M(v) - z_M^{\text{in}}(v))\Vert.
\label{gtthick}
\end{equation}

\begin{figure}[t]
	\noindent\resizebox{\textwidth}{!}{
	\begin{tikzpicture}
		\newcommand{\networkLayer}[6]{
			\def\a{#1} 
			\def\b{0.02}
			\def\c{#2} 
			\def\t{#3} 
			\def\d{#4} 

			\draw[line width=0.3mm](\c+\t,0,\d) -- (\c+\t,\a,\d) -- (\t,\a,\d);                                                      
			\draw[line width=0.3mm](\t,0,\a+\d) -- (\c+\t,0,\a+\d) node[midway,below] {#6} -- (\c+\t,\a,\a+\d) -- (\t,\a,\a+\d) -- (\t,0,\a+\d); 
			\draw[line width=0.3mm](\c+\t,0,\d) -- (\c+\t,0,\a+\d);
			\draw[line width=0.3mm](\c+\t,\a,\d) -- (\c+\t,\a,\a+\d);
			\draw[line width=0.3mm](\t,\a,\d) -- (\t,\a,\a+\d);

			\filldraw[#5] (\t+\b,\b,\a+\d) -- (\c+\t-\b,\b,\a+\d) -- (\c+\t-\b,\a-\b,\a+\d) -- (\t+\b,\a-\b,\a+\d) -- (\t+\b,\b,\a+\d); 
			\filldraw[#5] (\t+\b,\a,\a-\b+\d) -- (\c+\t-\b,\a,\a-\b+\d) -- (\c+\t-\b,\a,\b+\d) -- (\t+\b,\a,\b+\d);

			\ifthenelse {\equal{#5} {}}
			{} 
			{\filldraw[#5] (\c+\t,\b,\a-\b+\d) -- (\c+\t,\b,\b+\d) -- (\c+\t,\a-\b,\b+\d) -- (\c+\t,\a-\b,\a-\b+\d);} 
		}
		
		\tikzstyle{vecArrow} = [thick, decoration={markings,mark=at position1 with {\arrow[semithick]{open triangle 60}}},
   double distance=3pt, shorten >= 5.5pt,
   preaction = {decorate},
   postaction = {draw,line width=3pt, white,shorten >= 4.5pt}]

		\node at (-7.05,0.5){\begin{tabular}{l}\bf{Inner cortical}\\\bf{surface channel:}\end{tabular}};	
		\node at (-10,-2.5){\begin{tabular}{l}\bf{Outer cortical}\\\bf{surface channel:}\end{tabular}};	
		
		\draw [->] (-6,-7.5)  -- (12,-7.5);
		\node at (10.85,-7.75){\begin{tabular}{c}{Month \& Layer}\end{tabular}};
		
		\draw [] (-6,-7.5) -- (-6,-7.35);
                 \node at (-6,-7.75){\begin{tabular}{c}{1}\end{tabular}};
                 \draw [] (-3,-7.5) -- (-3,-7.35);
                 \node at (-3,-7.75){\begin{tabular}{c}{2}\end{tabular}};
                 \draw [] (0,-7.5) -- (0,-7.35);
                 \node at (0,-7.75){\begin{tabular}{c}{3}\end{tabular}};
                 \draw [] (4.5,-7.5) -- (4.5,-7.35);
                 \node at (4.5,-7.75){\begin{tabular}{c}{4}\end{tabular}};
                 \draw [] (6.25,-7.5) -- (6.25,-7.35);
                 \node at (6.25,-7.75){\begin{tabular}{c}{5}\end{tabular}};
                 \draw [] (8,-7.5) -- (8,-7.35);
                 \node at (8,-7.75){\begin{tabular}{c}{6}\end{tabular}};
                 
		\node[] (input image) at (-4.25,0.5) {\includegraphics[height=20mm]{figures/color_inner.png}};
		\draw [->,>=stealth] (-2.7,0.5) -- (-1.9,0.5);
		\draw [->,>=stealth] (-2.7,0.3) -- (-1.9,0.3);
		
		\node[] (input image) at (-7.25,-2.5) {\includegraphics[height=20mm]{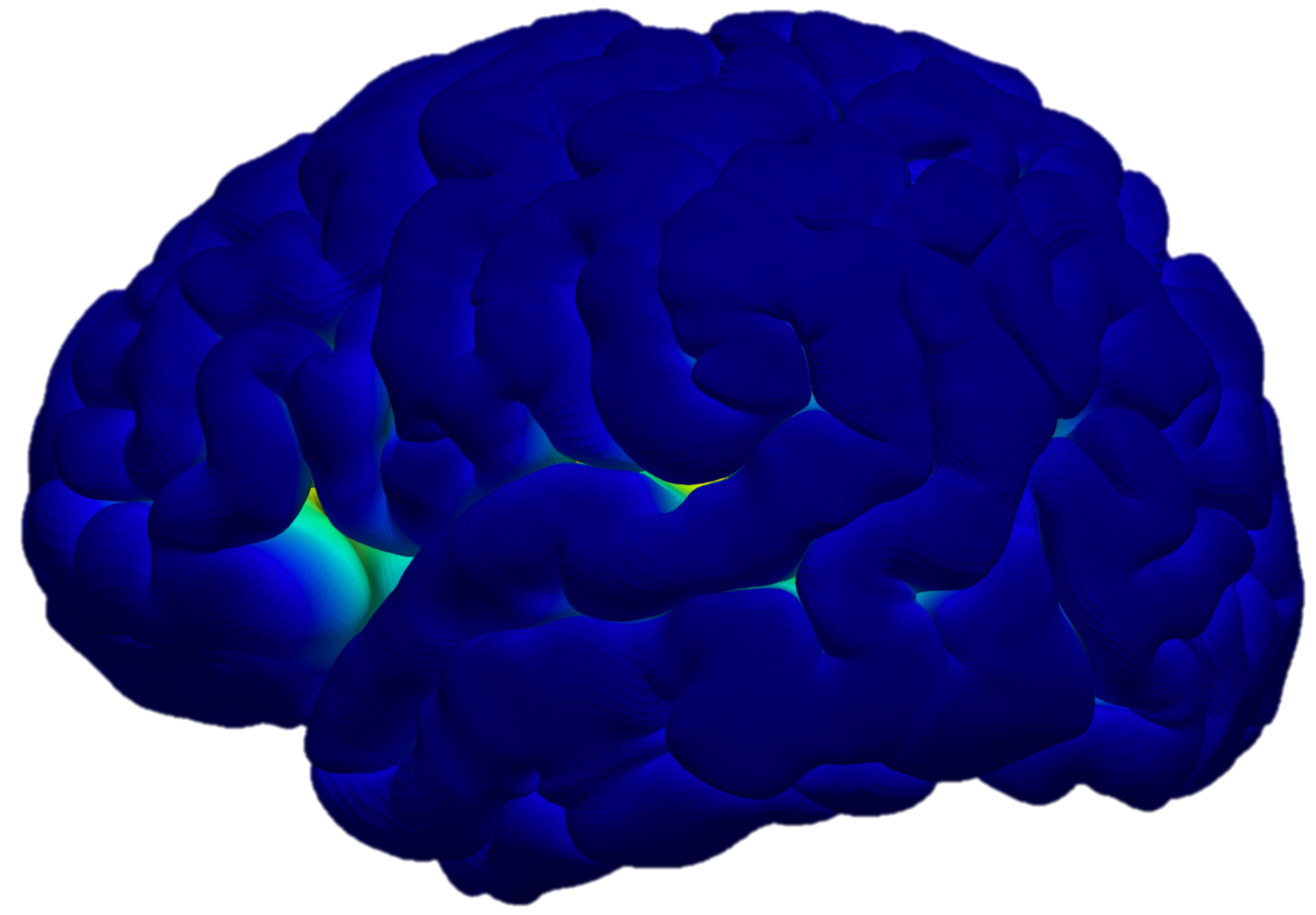}};
		\draw [->,>=stealth] (-5.75,-2.5) -- (-4.95,-2.5);
		\draw [->,>=stealth] (-5.75,-2.7) -- (-4.95,-2.7);
		
		\node at (-0.5,3.5){\begin{tabular}{c}$(|\mathcal{V}|\times 3)$\\ $\times 60$\end{tabular}};
		\node at (1.5,3.5){\begin{tabular}{c}$(|\mathcal{V}|\times 36)$\\$ \times 60$\end{tabular}};
		\node at (5,3.5){\begin{tabular}{c}$(|\mathcal{V}|\times 72)$\\ $\times 60$\end{tabular}};
		\node at (9,4){\begin{tabular}{c}$(|\mathcal{V}|\times 36)$\\$ \times 60$\end{tabular}};
		\node at (11.75,4.5){\begin{tabular}{c}$(|\mathcal{V}|\times 18)$\\$ \times 60$\end{tabular}};
		\node at (13.25,4.5){\begin{tabular}{c}$(|\mathcal{V}|\times 9)$\\$ \times 60$\end{tabular}};
	
		\node at (-4,-4.5){\begin{tabular}{c}$|\mathcal{V}|\times 36$\end{tabular}};
		\node at (-1.25,-4.5){\begin{tabular}{c}$|\mathcal{V}|\times 72$\end{tabular}};
		\node at (2.25,-4.5){\begin{tabular}{c}$|\mathcal{V}|\times 36$\end{tabular}};
		\node at (7.9,2.9){\begin{tabular}{c}$|\mathcal{V}|\times 3$\end{tabular}};
		\node at (5.75,-6.75){\begin{tabular}{c}$|\mathcal{V}|\times 18$\end{tabular}};
		\node at (7.25,-6.75){\begin{tabular}{c}$|\mathcal{V}|\times 9$\end{tabular}};
		\node at (8.5,-6.75){\begin{tabular}{c}$|\mathcal{V}|\times 3$\end{tabular}};
		\node at (14.5,4.25){\begin{tabular}{c}$|\mathcal{V}|\times 3$\end{tabular}};
		
		
		
		\filldraw[draw=black,fill=blue!10] (-11.25,6) rectangle (-9.45,5.8);
		\node at (-5.775,5.9){\begin{tabular}{l}Surface patch uniformization (Eq.~\eqref{feature}, Fig.~\ref{gcnn}(a))\end{tabular}};	
		
		\filldraw[draw=black,fill=gray!15] (-11.25,5.5) rectangle (-9.45,5.3);
		\node at (-6.45,5.4){\begin{tabular}{l}Spatial convolution (Eq.~\eqref{gconv}, Fig.~\ref{gcnn}(b))\end{tabular}};
		
		\filldraw[draw=black,fill=red!10] (-11.25,5) rectangle (-9.45,4.8);
		\node at (-6.95,4.9){\begin{tabular}{l}Fully convolutional output layer\end{tabular}};	
		
		\draw [-] (0,5.975) -- (0.5,5.975) -- (0.5,5.825);
		\draw [->,>=stealth] (0,5.825) -- (1,5.825);
		\node at (2.625,5.9){\begin{tabular}{l}Layer concatenation\end{tabular}};	
		
		\draw [->,>=stealth] (-0,5.475) -- (1,5.475);
		\draw [->,>=stealth] (-0,5.325) -- (1,5.325);
		\node at (2.225,5.4){\begin{tabular}{l}Input/Output\end{tabular}};	
		
		\draw [thick, dash dot] (0,-1.35) -- (-1.35,-1.35) -- (-1.35,2.75);
		
		\networkLayer{3.0}{0.1}{-0.5}{0.0}{color=blue!10}{}
		\networkLayer{3.0}{0.8}{-0.2}{0.0}{color=gray!15}{} 
		
		\draw [thick, dash dot] (-1.35,2.75) -- (0,2.75) -- (0,-1.35);
		\node at (-3.3,2.5){\begin{tabular}{c}Spatial graph\\convolutional \\layer (Fig. \ref{gcnn})\end{tabular}};		
		\draw [->,>=stealth, thick, dash dot]  (-2.2,2.5) -- (-1.5,2.5);
		\networkLayer{3.0}{0.1}{-0.5}{8.0}{color=blue!10}{}
		\networkLayer{3.0}{0.8}{-0.2}{8.0}{color=gray!15}{} 
		
		\networkLayer{3.0}{0.8}{1.2}{0.0}{color=blue!10}{}
		\networkLayer{3.0}{1.5}{2.2}{0.0}{color=gray!15}{}
		
		\networkLayer{3.0}{0.8}{1.2}{8.0}{color=blue!10}{}
		\networkLayer{3.0}{1.5}{2.2}{8.0}{color=gray!15}{}
		
		\networkLayer{3.0}{1.5}{4.3}{0.0}{color=blue!10}{}
		\networkLayer{3.0}{0.8}{6}{0.0}{color=gray!15}{}
		
		\draw [->,>=stealth] (6.25,1.1) -- (8.65,1.8);
		
		\networkLayer{3.0}{1.5}{4.3}{8.0}{color=blue!10}{}
		\networkLayer{3.0}{0.8}{6}{8.0}{color=gray!15}{}
		
		\networkLayer{3.0}{0.8}{9}{-2.5}{color=blue!10}{}
		\networkLayer{3.0}{0.4}{10}{-2.5}{color=gray!15}{}
		
		\networkLayer{3.0}{0.1}{8.25}{7}{color=red!10}{}
		\draw [->,>=stealth] (3.2,-2.15) -- (4.25,-4);
		\draw [-] (5.15,-1.85) -- (5.4,-1.85) -- (3.9,-3.4);

		\networkLayer{3.0}{0.1}{8.25}{1}{color=red!10}{}
		
		\draw [->,>=stealth] (5.15,-1.75) -- (5.95,-1.75);
		\draw [->,>=stealth] (5.15,-1.95) -- (5.95,-1.95);
		
		\draw [-] (7.4,0.65) -- (7.65,0.65) -- (8.38,1.72);
		
	        \node[] (output image) at (7.5,-1.75) {\includegraphics[height=20mm]{figures/color_outer.png}};
		
		\networkLayer{3.0}{0.8}{11}{14}{color=blue!10}{}
		\networkLayer{3.0}{0.4}{12}{14}{color=gray!15}{}
		
		
		\networkLayer{3.0}{0.4}{11}{-2.5}{color=blue!10}{}
		\networkLayer{3.0}{0.2}{11.6}{-2.5}{color=gray!15}{}
		
		\networkLayer{3.0}{0.4}{13}{14}{color=blue!10}{}
		\networkLayer{3.0}{0.2}{13.6}{14}{color=gray!15}{}
		
		\networkLayer{3.0}{0.2}{12.4}{-2.5}{color=blue!10}{}
		\networkLayer{3.0}{0.1}{12.8}{-2.5}{color=gray!15}{}
		
		\networkLayer{3.0}{0.2}{14.4}{14}{color=blue!10}{}
		\networkLayer{3.0}{0.1}{14.8}{14}{color=gray!15}{}
		
		\draw [->,>=stealth] (7.4,0.75) -- (8.2,0.75);
		\draw [->,>=stealth] (7.4,0.55) -- (8.2,0.55);
		\node[] (output image) at (9.75,0.75) {\includegraphics[height=20mm]{figures/color_inner.png}};

		\networkLayer{3.0}{0.1}{13.5}{-2.5}{color=red!10}{}
		
		\networkLayer{3.0}{0.1}{15.5}{14}{color=red!10}{}
		
		\draw [->,>=stealth] (14,2) -- (14.8,2);
		\draw [->,>=stealth] (14,1.8) -- (14.8,1.8);
		\node[] (output image) at (16.4,2) {\includegraphics[height=20mm]{figures/color_inner.png}};
		
		\draw [->,>=stealth] (9.65,-4.3) -- (10.45,-4.3);
		\draw [->,>=stealth] (9.65,-4.5) -- (10.45,-4.5);
		\node[] (output image) at (12,-4.3) {\includegraphics[height=20mm]{figures/color_outer.png}};
		
	\end{tikzpicture}
	}
	\caption{GCNN longitudinal prediction network. Each graph convolutional layer consists of a surface patch uniformization sub-layer (blue blocks) with 60 learnable Gaussian kernels, followed by a spatial convolution sub-layer (gray blocks) with feature map sizes of 36, 72, 36, 18, 9, 3. Note that the surface uniformization sub-layers do not change feature map size.}
	\label{fw}
\end{figure}

\subsubsection{Network Architecture.}
\label{3.3.2}
The proposed longitudinal prediction network has two independent prediction channels with the same architecture to cater to the inner and outer cortical surfaces (see Fig.~\ref{fw}). 
The depth of the network increases with time, allowing more complex mappings to be learned as prediction further in time needs to be carried out. 
At each target time point $M$ (i.e., $3^{\text{rd}}$, $6^{\text{th}}$ month), the network outputs the predicted cortical surface displacements (i.e., $O^{\text{in}}_M, O^{\text{out}}_M$ in  Eqs.~\eqref{ffi1}--\eqref{ffi2}). 
Additionally, at the $3^{\text{rd}}$ month, the hidden feature maps from the $3^{\text{rd}}$ spatial graph convolution sub-layer will be concatenated with the predicted surface displacements at $3^{\text{rd}}$ month to be fed together into the next ($4^{\text{th}}$) surface patch uniformization sub-layer for prediction at the $6^{\text{th}}$ month. 
Layer concatenation as described above can be interpreted as gathering of dynamic growth momentum (hidden features at the $3^{\text{rd}}$ month) and static status (cortical geometry at the $3^{\text{rd}}$ month). During training, the cortical growth across all time points are learned as a whole, promoting the temporal consistency of longitudinal predictions.
The network architecture is generic and can be modified to take into consideration more time points by increasing network depth and adding more output and concatenation layers. 


\subsubsection{Loss functions.}
\label{3.3.3}

We employ $L^2$-norm, denoted as $\text{Loss}_{\text{displacement}}^M$, to measure the errors in the predicted vertex displacements $O_M^{\text{in}}$ and $O_M^{\text{out}}$ of the inner and outer cortical surfaces at the $M$-th month ($M = 3, 6$). We also include a cortical thickness loss term to constrain the spatial consistency between the predicted $\mathcal{G}^{\text{in}}_M$ and $\mathcal{G}^{\text{out}}_M$:
\begin{equation}
\text{Loss}_{\text{thickness}}^M = \sum_{v\in\mathcal{V}}|\text{thickness}^{\text{p}}(v) - \text{thickness}^{\text{g}}(v)|,
\end{equation}
where $\text{thickness}^{\text{p}}(\cdot)$ and $\text{thickness}^{\text{g}}(\cdot)$ are the predicted and ground-truth thickness, respectively.

To fully utilize the incomplete data (see Table~\ref{info}), we use a binary flag to indicate the nonexistence (flag `0' ) or existence (flag `1') of a ground-truth surface. The flag is used to ensure that the loss associated with a nonexistent surface is not contributing to back-propagation. More specifically, this is achieved by defining the total loss of the network as
\begin{equation}
\text{Loss} = \sum_{M=3,6} \text{flag}^M \cdot (\alpha\cdot\text{Loss}^M_{\text{displacement}} + (1-\alpha)\cdot\text{Loss}^M_{\text{thickness}}).
\end{equation}
where $\alpha$ ($0<\alpha<1$) controls the relative contributions of $\text{Loss}^M_{\text{displacement}}$ and $\text{Loss}^M_{\text{thickness}}$.
As a consequence, our network is flexible and does not require longitudinal data that are complete at all  $1^{\text{st}}, 3^{\text{rd}}$ and $6^{\text{th}}$ months for training. Each available ground-truth cortical surface can contribute to the learning of the growth trajectories.

\section{Experiments}
\label{4}
We compared the proposed longitudinal prediction network (GCNN-LP) with two methods: (1) Affine transformation (AF) \cite{kanatani1996affine}; (2) Independent prediction networks (GCNN-IPs). Specifically, the GCNN-IP for surface prediction at the $3^{\text{rd}}$ month is implemented by removing all layers after the output layer for the $3^{\text{rd}}$ month in GCNN-LP. GCNN-IP for the $6^{\text{th}}$ month is implemented by removing the output and concatenation layers of GCNN-LP at the $3^{\text{rd}}$ month. All inputs and outputs of GCNN-IPs are the same as those defined in GCNN-LP.

We chose 3 infant subjects (see Table~\ref{info}) with data available at $1^{\text{st}}$,$3^{\text{rd}}$ and $6^{\text{th}}$ months for testing. Other subjects were used for training. For each surface mesh $\mathcal{G}$, in total the number of vertices $|\mathcal{V}|=10242$ and the number of triangular faces $|\mathcal{F}|=20480$. Local geodesic polar grids with radius $\rho_{\mathcal{D}}=2\,\text{mm}$ were constructed on the baseline inner and outer cortical surfaces. Each grid was partitioned into 60 bins ($N_{\rho}=5,N_{\theta}=12$), as shown in Fig.~\ref{grid}. The corresponding 60 learnable Gaussian kernels (Eq.~\eqref{gaussian}) were initialized with random means and variances. Training for each model was performed with 25k maximum updates using Adam, $10^{-4}$ learning rate for the first 5k iterations, and $10^{-5}$ for the remaining iterations. 

To quantitatively evaluate the prediction results, we computed the median absolute error (MAE) for both cortical surface (cs) and cortical thickness (ct) predictions:
\begin{align}
\text{MAE}_{\text{cs}} & = \text{median}_{v\in\mathcal{V}}\{\Vert(x^{\text{p}}(v)-x^{\text{g}}(v), y^{\text{p}}(v)-y^{\text{g}}(v), z^{\text{p}}(v)-z^{\text{g}}(v))\Vert\},\\
\text{MAE}_{\text{ct}}  & =  \text{median}_{v\in\mathcal{V}}\{|\text{thickness}^{\text{p}}(v) - \text{thickness}^{\text{g}}(v)|\}
\end{align}
where `p' and `g' denote the prediction and ground-truth, respectively.

Table~\ref{stat} summarizes the prediction performance of the different methods. 
Compared with AF, the better performance of all GCNN models in predicting cortical surfaces indicates that GCNNs are capable of learning the non-linearity in cortical growth, which cannot be captured by AF. The smaller errors achieved by GCNNs over AF in cortical thickness prediction validate the spatial consistency between the inner and outer cortical surfaces predicted by GCNNs. 
GCNN-LP yields greater accuracy than GCNN-IP in all prediction tasks.

\begin{figure}[h]
\centering
\includegraphics[width = \textwidth]{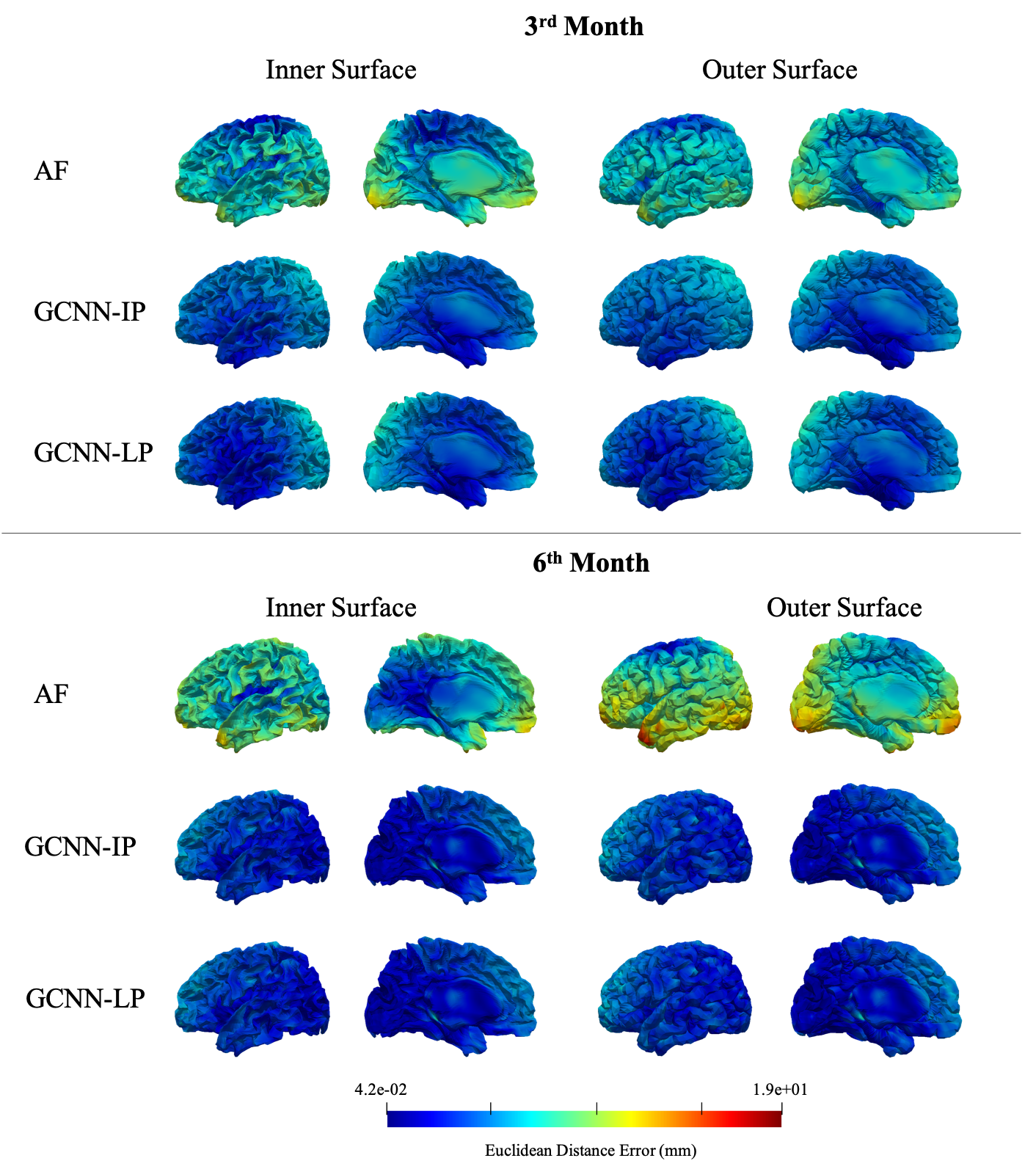}
\caption{Surface prediction errors.}
\label{CS}
\end{figure}

\begin{figure}[h]
\centering
\includegraphics[width = 0.9\textwidth]{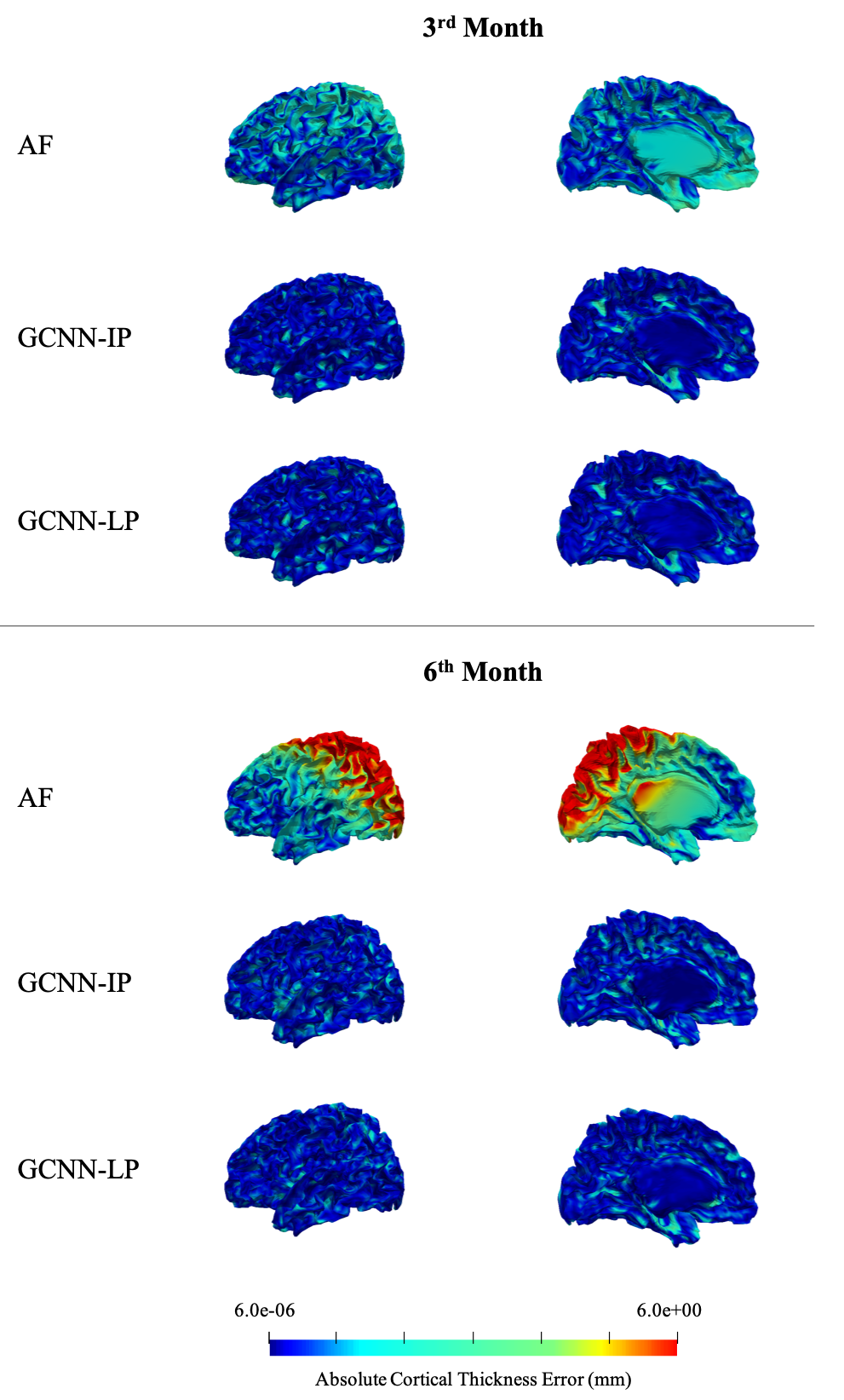}
\caption{Errors in cortical thickness computed from the predicted cortical surfaces.}
\label{CT}
\end{figure}

Fig.~\ref{CS} shows the vertex-wise surface prediction errors in terms of $L^2$ Euclidean distances of corresponding vertices on the inner and outer surfacess.
The corresponding error maps for cortical thickness computed from the predicted cortical surfaces are shown in Fig.~\ref{CT}. Clearly, GCNN-LP and GCNN-IP yield much smaller prediction errors than AF across different cortical regions.
%

\begin{table}[t]
\caption{Quantitative comparison of AF, GCNN-IP, and GCNN-LP at the $3^{\text{rd}}$ and $6^{\text{th}}$ months using median absolute errors (MAE) averaged across testing subjects.} 
\begin{center}
    \begin{tabular}{ccP{1.5cm}P{1.5cm}P{1.5cm}P{1.5cm}P{1.5cm}P{1.5cm}}  
       \toprule 
      \multirow{2}{*}{ \thead{Month}} &
\multirow{2}{*}{ \thead{Method}} &
  \multicolumn{2}{c}{\thead{~~Cortical Thickness~~}} &
  \multicolumn{2}{c}{\thead{~~Inner Surface~~}}  &
  \multicolumn{2}{c}{\thead{~~Outer Surface~~}} \\
\cmidrule(lr){3-4} \cmidrule(lr){5-8}  
  &  &\multicolumn{2}{c}{$\text{MAE}_{\text{ct}}$ ($\text{mm}$)}  & \multicolumn{4}{c}{$\text{MAE}_{\text{cs}}$ ($\text{mm}$)} \\
\midrule
      \multirow{3}{*}{3} & AF & 
       \multicolumn{2}{c}{1.0872} &
      \multicolumn{2}{c}{6.0268} & 
      \multicolumn{2}{c}{5.2221} \\
      & GCNN-IP & 
      \multicolumn{2}{c}{0.3146} &
      \multicolumn{2}{c}{4.3052} & 
      \multicolumn{2}{c}{4.2819} \\
      & {\bf{GCNN-LP}} &  
      \multicolumn{2}{c}{\bf{0.3085}}  &
      \multicolumn{2}{c}{\bf{3.7780}} & 
      \multicolumn{2}{c}{\bf{3.8380}}  \\
     \hline
      \multirow{3}{*}{6} & AF & 
     \multicolumn{2}{c}{3.0784} &
     \multicolumn{2}{c}{6.7232} & 
     \multicolumn{2}{c}{7.2481} \\       
     & GCNN-IP & 
     \multicolumn{2}{c}{0.3934} &
     \multicolumn{2}{c}{3.0116} & 
     \multicolumn{2}{c}{3.1379} \\
     & {\bf{GCNN-LP}} & 
     \multicolumn{2}{c}{\bf{0.3560}}  &
     \multicolumn{2}{c}{\bf{2.9585}}  & 
     \multicolumn{2}{c}{\bf{3.0032}}  \\
     \bottomrule
    \end{tabular}
\end{center}
\label{stat}
\end{table}

\section{Conclusion}

We propose a novel longitudinally consistent prediction method, based on spatial graph convolutional neural network, for the prediction of infant cortical surfaces. Our method is able to learn intrinsic, non-linear growth features via spatial convolution directly applied on the cortical surfaces. Predicting the cortical growth across all target time points within a single network, the proposed method models the growth trajectories with temporal consistency. Experimental results demonstrated that our method is capable of capturing the growth trajectories for accurate longitudinal surface prediction. 
\nocite{*}
 \bibliographystyle{splncs04}
 \bibliography{sub/references}
\end{document}